\documentclass[11pt,a4paper]{article}
\usepackage{naaclhlt2018}
\usepackage{times}
\usepackage{latexsym}
\usepackage{graphicx}

\usepackage{url}

\usepackage{amsmath}
\DeclareMathOperator*{\argmax}{argmax}
\usepackage[utf8x]{inputenc}
\usepackage{subcaption}

\aclfinalcopy 

\title{Duluth UROP at SemEval-2018 Task 2: Multilingual Emoji Prediction with Ensemble Learning and Oversampling}
\author{Shuning Jin \and Ted Pedersen \\
Department of Computer Science \\
University of Minnesota\\
Duluth, MN 55812 USA \\
{\tt \{jinxx596,tpederse\}@d.umn.edu} \\
{\tt https://github.com/shuningjin/SemEval2018-Task2-EmojiDetection}}

\date{}

\begin{document}
\maketitle
\begin{abstract}
This paper describes the Duluth UROP systems that participated
in SemEval--2018 Task 2, Multilingual Emoji Prediction. We
relied on a variety of ensembles made up of classifiers 
using Naive Bayes,
Logistic Regression, and Random Forests. We used unigram and
bigram features and tried to offset the skewness of the data
through the use of oversampling. Our task evaluation 
results place us 19th of 48
systems in the English evaluation, and 5th of 21 in the Spanish.
After the evaluation we realized that some simple changes to preprocessing could significantly improve our results. After
making these changes we attained results that would have placed
us sixth in the English evaluation, and second in the Spanish.
\end{abstract}

\section{Introduction}

Emoji prediction of tweets is an emerging problem \cite{Saggion17}
which combines the nuances of sentiment analysis with the noisy data
characteristic of social media. SemEval--2018 Task 2
\cite{semeval2018task2} adds to this the challenge of multilingual processing,
where both Spanish and English tweets are used. 

Given the relatively large amount of training data available for the
task (see Section 2) we decided to approach this as a classification
task, where we used relatively simple unigram and bigram features
in combination with standard machine learning algorithms. We
particularly focused on the use of Multinomial Naive Bayes,
Logistic Regression, and Random Forests, all of which were provided
to us via the scikit learn package \cite{scikit-learn}. Given the challenging
nature of this task we decided to employ a variety of ensemble approaches, 
since no single classifier seemed likely to perform well in all cases.

In the sections that follow we summarize the experimental data
used in the task, and then describe the methods we employed for
supervised learning, ensemble building, and oversampling. We close
by interpreting and discussing our results.

\section{Experimental Data}

The task organizers created both training and test
data made up of Spanish and English tweets (separately).
The training data consists of 100,000 Spanish tweets
and 500,000 English tweets. The test data is made up of
10,000 Spanish tweets and 50,000 English tweets.
Each instance consists of a single tweet, where
19 different emojis were observed in the Spanish
data, and 20 emojis were observed in the English.

We collected the training tweets via the Twitter API
on November 10--11, 2017.
By that time some of the tweets selected by the
organizers had been deleted or made private, but
we were still able to download the vast majority
of training tweets. For English we downloaded 490,272
tweets, so 9,628 were unavailable. For Spanish we
obtained 98,657 tweets, so 1,343 were unavailable.

One of the most striking aspects of this data for
both English and Spanish is that the number of
classes (emojis) is relatively large (20 and 19
respectively), and that the distribution of these
classes is skewed. In the English training data
the most common emoji is  {\includegraphics[height=0.30cm,width=0.30cm]{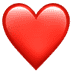}} which
occurs 21.7\% of the time. The next most frequent
emoji is {\includegraphics[height=0.33cm,width=0.33cm]{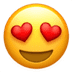}}
which occurs 10.5\% of the time. The two emojis
were also the most frequent in the Spanish data,
occurring 19.9\% and 13.7\% of the time. By contrast
16 of the emojis occurred less than 5\% of the time
in English, and 14 occurred less than 5\% in the
Spanish.

The evaluation measure used to rank systems in
this task is the  F--measure, which rewards
systems that are able to predict instances in the
low frequency classes correctly. Given that we
decided to employ oversampling in order to try
to improve our results on the low frequency classes
which had the negative effect of degrading our
performance on the more frequent classes.

\section{System Description}
\subsection{Preprocessing}

First, the text is normalized to lowercase. In preliminary
experiments, we find that removing all punctuations 
reduces performance, thus we decided to only remove commas. 
In our task evaluation experiments we removed all non-ASCII
characters, but then post--evaluation decided to keep most
of them. Then, we tokenize the tweets with NLTK word 
tokenizer and identify unigrams and bigrams as potential features.
To reduce noise, a document frequency cutoff of 5 is applied to sift out 
unigram and bigram features that occur in at least five tweets. 

When applied to the English data, this process 
results in 166,681 features, including 35,197 unigrams and 
131,484 bigrams. The Spanish data is made up of 40,420 features, 
including 12,356 unigrams and 28,061 bigrams. For text 
representation, we adopt the bag-of-words model. Each tweet is converted to a vector based on n-grams by Count Vectorizer.

\subsection{Oversampling}

Faced with the skewness of both the English and Spanish tweets,
we introduce oversampling 
to adjust the class distribution to reduce bias. We use 
Synthetic Minority Oversampling Technique (SMOTE), where 
the minority classes are oversampled by creating synthetic 
examples using k nearest neighbors. 
We use imblearn in scikit-learn library for oversampling. 

In our case, all classes are resampled to have equal number with the majority 
class \includegraphics[height=0.30cm,width=0.30cm]{img/red_heart.png}. As a result, the resampled Spanish data has a 
size of 373,825, with class size of 19,675, and English has a size of 2,130,180, with 
class size of 106,509.

\subsection{Base Classifier}

We use scikit-learn library for the base classifiers and 
the first-level ensemble classifier below. And the 
second-level ensemble is constructed with mlxtend library, 
which is compatible with scikit-learn.

\subsubsection{Multinomial Naive Bayes (MNB):}

MNB is a probabilistic classifier based on 
integer feature counts. It is simple yet 
powerful for text classification, especially for short 
documents \cite{Wang2012}. To eliminate zero 
counts, we use additive smoothing with a parameter value of 0.5.


\subsubsection{Logistic Regression (LR):}

While NB assumes strong conditional independence, LR is 
more robust to correlated features. We use a LR with L2 
regularization to reduce overfitting. It uses the one-vs-rest 
(OvR) scheme for multiclass classification.

\subsubsection{Random Forest (RF):}

As an enhancement of decision tree, we use the 
RF classifier, which ensembles a multitude of decision 
trees. By fitting on sub-samples of the dataset, 
RF improves accuracy and reduces overfitting by averaging. 
We use 20 trees here.

\subsection{Ensemble Classifier}

We build an ensemble classifier to combine the strengths 
of a collection of base models. The ensemble method is soft 
voting, where the calibrated member classifiers cast weighted 
votes for classes based on predicted probabilities. The ensemble is also a calibrated classifier, who can either predict associated 
probabilities based on weighted sum, or a class with maximum probability.

\begin{equation}
P(c_j) =\sum_{i\in ensemble}w_iP_i(c_j)
\end{equation}

\begin{equation}
c = \argmax_{c_j} P(c_j)
\end{equation}

Our ensemble has two levels. On the base level, 
we include a diverse collection of heterogeneous classifiers: 
MNB, LR, RF, with 
weights (1.1,1,1) for Spanish and (1.5,6,1) for English.


On the second level, we train the base ensemble respectively 
on the original task data (Ensemble1) and oversampled data (Ensemble2). On the one hand, oversampling can adjust class distribution so that rare classes are well represented. On the other hand, it may exacerbate overfitting problem in the context of noisy data, and consequently harms accuracy. To seek a balance, we devise a meta-ensemble classifier (Meta) including both Ensemble1 and Ensemble2, with weights 
(3,1) for Spanish and (4,1) for English.


The weights for the ensembles were set in a non-systematic
fashion via trial and error. In future work, we would like to 
arrive at these weights in a more rigorous fashion.

\subsection{Evaluation Metric}
For individual classes, $F_1$ score is calculated as:
$$ F = \frac{2Precision \cdot Recall}{Precision+Recall} $$
\noindent
The overall classification performance of the system is measured by macro-averaged $F_1$ score:
$$ F_{macro} = \frac{1}{k} \sum_{i=1}^k F_i$$
where k is total number of classes.

\section{Results and Discussion}

\subsection{Task Evaluation}
For task evaluation, our two submitted systems are: Ensemble1 and Meta.
The official results are shown for English in Table \ref{table:English1} and Spanish in Table \ref{table:Spanish1}. Our overall F-score was competitive in both the English (19th of 48) and Spanish tasks (5th of 21).

\begin{table}
\centering
\begin{tabular}{|c|ccc|c|}\hline
\textbf{Method} & \textbf{F1} & \textbf{P} & \textbf{R} & \textbf{Acc.}  \\ \hline
Ensemble1 & 26.37 & 28.10& 27.41& 34.43 \\ \hline
Meta & 26.59 & 27.18 & 27.87 & 33.80 \\ \hline
\end{tabular}
\caption{English Task Evaluation Results}
\label{table:English1}
\end{table}

\begin{table}
\centering
\begin{tabular}{|c|ccc|c|}\hline
\textbf{Method} & \textbf{F1} & \textbf{P} & \textbf{R} & \textbf{Acc.}  \\ \hline
Ensemble1 & 16.59 & 18.03& 17.84& 29.67 \\ \hline
Meta & 16.75 & 17.11 & 18.10& 28.51 \\ \hline
\end{tabular}
\caption{Spanish Task Evaluation Results}
\label{table:Spanish1} 
\end{table}

\subsection{Post Evaluation}

In post-evaluation experiments, we revised the preprocessing by
including most non-ASCII characters and modified the weights
assigned for ensembles. As a result, the system performance improved greatly, which was largely attributed to the changes in preprocessing.

\begin{figure}
	\includegraphics[width=\linewidth]{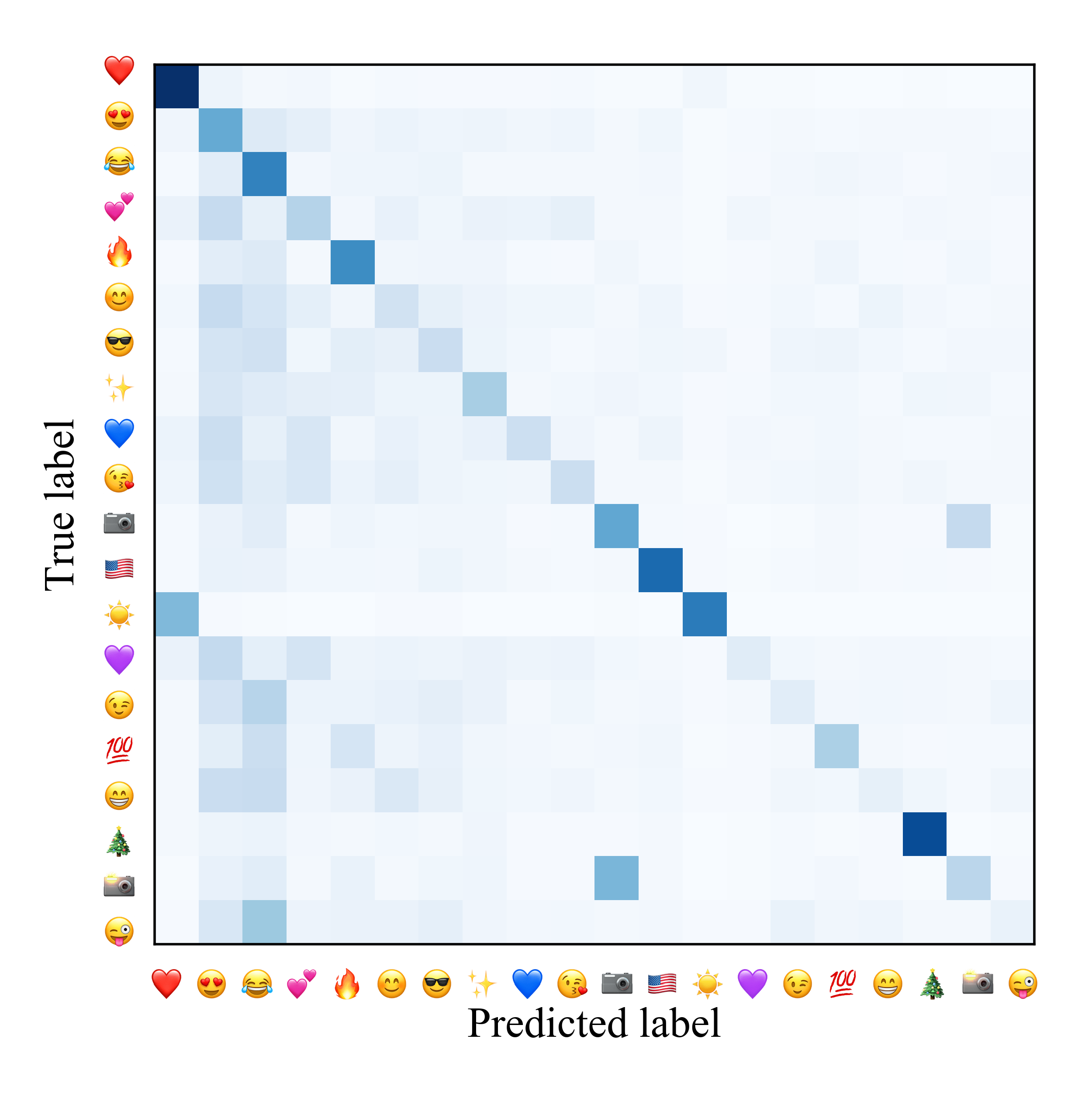}
	\caption{English Meta Confusion Matrix}
	\label{figure:EnglishCM}
\end{figure}

\begin{figure}
	\includegraphics[width=\linewidth]{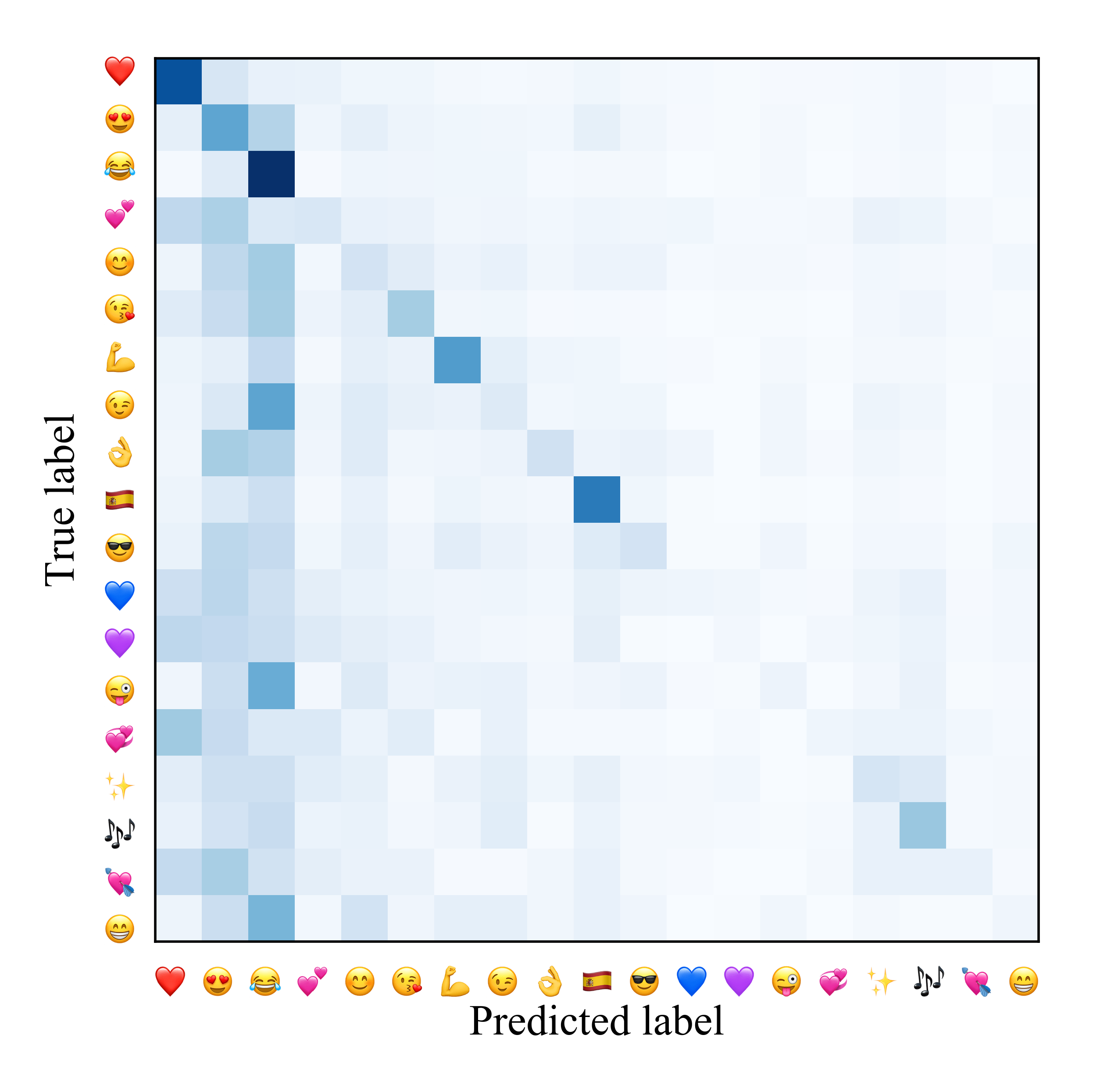}
	\caption{Spanish Meta Confusion Matrix}
	\label{figure:SpanishCM}
\end{figure}

The system was trained on the whole training data, and tested with the official test data. We show post-evaluation results for English in Table \ref{table:English2} and
for Spanish in Table \ref{table:Spanish2}. The confusion matrices of Meta classifier are shown in Figure \ref{figure:EnglishCM} and Figure \ref{figure:SpanishCM}.

\begin{table}
\centering
\begin{tabular}{|c|ccc|c|} \hline
\textbf{Method} & \textbf{F1} & \textbf{P} & \textbf{R} & \textbf{Acc.}  \\ \hline
  MNB-P & 30.21 &30.78 &31.58 & 42.22 \\ 
  LR-P & 32.73 & 35.05 & 32.08 &44.79 \\ 
  RF-P & 24.49 & 30.13 & 24.41 & 39.01 \\ \hline
Ensemble1-P & 33.03 & 34.68 &33.09 & 45.68 \\ 
Ensemble2-P & 31.85 & 31.38& 33.14 & 42.08 \\ \hline
Meta-P & 33.31 & 34.14& 33.61 & 45.58 \\ \hline
\end{tabular}
\caption{English Post-Evaluation Results}
\label{table:English2}
\end{table}

\begin{table}
\centering
\begin{tabular}{|c|ccc|c|}\hline
\textbf{Method} & \textbf{F1} & \textbf{P} & \textbf{R} & \textbf{Acc.}  \\ \hline
  MNB-P & 19.26 &19.92 &20.51 &35.07 \\ 
  LR-P & 18.43 & 20.98& 18.28&35.23 \\ 
  RF-P & 13.41 & 19.47& 13.78 &32.68 \\ \hline
Ensemble1-P & 19.58 & 21.13& 20.51& 37.05 \\ 
Ensemble2-P & 20.34 & 20.44& 21.55& 33.64 \\ \hline
Meta-P & 20.21 & 21.23 & 21.12& 36.74 \\ \hline
\end{tabular}
\caption{Spanish Post-Evaluation Results}
\label{table:Spanish2}
\end{table}

In the task evaluation we eliminated all non-ASCII characters during
preprocessing. After the evaluation period we realized that this
resulted in a significant loss
of accuracy. We revised our preprocessing for our post-evaluation
experiments and only removed
the following non-ASCII characters (shown as Unicode value, description):
(U+00B7, middle dot),
(U+2019, right single quotation mark),
(U+2018, left single quotation mark),
(U+2022, bullet),
(U+2026, horizontal ellipsis), and
(U+30FB, katakana middle dot).

Preserving non-ASCIIs is
important for both languages. Spanish using MNB has a F-score
of 16.77 without non-ASCII
and 19.08 when preserving all, and English has 25.47 without non-ASCII
and 30.00 when including. While their importance for Spanish is apparent
as accents are ubiquitous in Latin languages, their function
for English is relatively vague. Nevertheless, they are clearly providing
useful information in the tweets.

\subsection{Discussion}
In this section, we will discuss the results based on post-evaluation.

Ideally, we would hope an ensemble would outperform all of its 
components. Its performance counts on the accuracy and 
diversity of the members. While NB and LR are linear classifiers, RF is nonlinear. Also, the actual performance is sensitive 
to the assigned voting weights. Initially, we get a rough 
estimation based on individual performances, especially 
accuracy. For both English and Spanish, RF has a notable 
inferior score, so lower weight is expected. Then we 
manually perform experiments to find better weights. Due 
to the large size of data, it is computationally 
expensive to perform grid search. In the future, we would 
like to investigate if other automated methods could find 
optimal hyper-parameters more efficiently.

By fitting the ensemble with oversampled data, the overall accuracy drops.
However, the rare classes originally with low F-scores gain an increase.
This is desirable as we attempt to maximize the overall classification performance
for all classes, which is measured by macro-averaged F. In considering the weights in Meta, we perceive Ensemble1 a more reliable source as it shows
notably higher accuracy. Meta tends to outperform Ensemble1 in F-score.

It is worth mentioning that the baseline 
classifiers like MNB and LR have robust performance compared 
to other more complex systems. This suggest that improvement 
from ensembles may be limited for this challenging problem, and 
new perspectives are necessary.

Additionally, there are some interesting observations from the confusion matrices. For English tweets, 
\includegraphics[height=0.30cm,width=0.30cm]{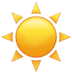}
is frequently misclassified as 
\includegraphics[height=0.30cm,width=0.30cm]{img/red_heart.png}. And there is apparent confusion between
\includegraphics[height=0.30cm,width=0.30cm]{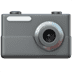}
and
\includegraphics[height=0.30cm,width=0.30cm]{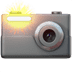}. For Spanish tweets, \includegraphics[height=0.30cm,width=0.30cm]{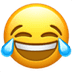} has highest accuracy. Meanwhile, many other emojis are misclassified to this label, typical ones including 
\includegraphics[height=0.30cm,width=0.30cm]{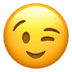} ,
\includegraphics[height=0.30cm,width=0.30cm]{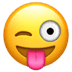} , and
\includegraphics[height=0.30cm,width=0.30cm]{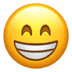} .

\section*{Acknowledgments}

The first author is grateful for the support of the 
Undergraduate Research Opportunity Program at the 
University of Minnesota.

\bibliography{naaclhlt2018}
\bibliographystyle{acl_natbib}

\end{document}